\documentclass{article}
\usepackage{ijcai24}
\usepackage{amssymb}
\usepackage{times}
\usepackage{soul}
\usepackage{url}
\usepackage[hidelinks]{hyperref}
\usepackage[utf8]{inputenc}
\usepackage[small]{caption}
\usepackage{graphicx}
\usepackage{subfigure}
\usepackage{amsmath}
\usepackage{amsthm}
\usepackage{booktabs}
\usepackage[switch]{lineno}
\usepackage{multirow}
\usepackage{indentfirst}
\usepackage{float}
\usepackage{algorithm}  
\usepackage{algorithmic}  

\newtheorem{definition}{Definition}

\title{DST-GTN: Dynamic Spatio-Temporal Graph Transformer Network for Traffic Forecasting}

\author{
Songtao Huang$^1$\and
Hongjin Song$^3$\and
Tianqi Jiang$^{4}$\and
Akbar Telikani$^{5}$\and
Jun Shen$^{5}$\and
Qingguo Zhou$^{1}$\and
Binbin Yong$^{1,*}$\And
Qiang Wu$^{2,}$\thanks{Corresponding author}\\
\affiliations
$^1$Lanzhou University\\
$^2$University of Electronic Science and Technology of China\\
$^3$Northwestern University\\
$^4$South China University of Technology\\
$^5$University of Wollongong\\
\emails
\{huangst21, zhouqg, yongbb\}@lzu.edu.cn,
hongjinsong2024@u.northwestern.edu,
ctjiangtianqi@mail.scut.edu.cn,
at952@uowmail.edu.au,
jshen@uow.edu.au,
qiang.wu@uestc.edu.cn
}

\begin{document}

\maketitle

\begin{abstract}
Accurate traffic forecasting is essential for effective urban planning and congestion management. 
Deep learning (DL) approaches have gained colossal success in traffic forecasting but still face challenges in capturing the intricacies of traffic dynamics.
In this paper, we identify and address this challenges by emphasizing that spatial features are inherently dynamic and change over time.
%
A novel in-depth feature representation, called Dynamic Spatio-Temporal (Dyn-ST) features, is introduced, which encapsulates spatial characteristics across varying times.
Moreover, a Dynamic Spatio-Temporal Graph Transformer Network (DST-GTN) is proposed by capturing Dyn-ST features and other dynamic adjacency relations between intersections.
The DST-GTN can model dynamic ST relationships between nodes accurately and refine the representation of global and local ST characteristics by adopting adaptive weights in low-pass and all-pass filters, enabling the extraction of Dyn-ST features from traffic time-series data.
Through numerical experiments on public datasets, the DST-GTN achieves state-of-the-art performance for a range of traffic forecasting tasks and demonstrates enhanced stability. 
%
%
\end{abstract}

\section{Introduction}

%
With the rapid development of urbanization and the increasing number of vehicles, accurate traffic forecasting can aid in managing numerous traffic-related issues, such as congestion, travel duration, and flow control~\cite{JinDiandShi2023AAAI}
%
Traditional traffic forecasting studies are mainly categorized into knowledge-driven and data-driven methods~\cite{ShaoZhang2022ProcVLDBEndow}. 
Knowledge-driven methods, including queuing and traffic flow theories, simulate drivers' behavior\cite{cascetta2013transportation}. 
However, they often overlook real-world complexities and the unpredictability of factors like road incidents or sudden weather shifts.
Data-driven methods, such as the autoregressive integrated moving average (ARIMA)~\cite{Williams2003JournalofTransportationEnginering} and the vector autoregressive model (VAR)~\cite{lu2016integrating}, generally treat traffic forecasting as a straightforward time series issue. 
These methods struggle with the high nonlinearity of traffic data because they rely on stationary assumptions, which are rarely met in actual traffic conditions.

The advent of deep learning (DL)~\cite{lecun2015Nature} enabled the intricate capture of spatial-temporal correlations in traffic flows~\cite{lv2014IEEE,li2018diffusion}. 
Earlier DL-based approaches transformed traffic data into spatial-temporal (ST) grids, employing Convolutional Neural Networks (CNNs) to capture spatial dependencies and Recurrent Neural Networks (RNNs) for temporal dependencies~\cite{Yao2018AAAI}.
Recent DL studies focused on Spatio-Temporal Neural Graph Networks (STGNNs), which effectively utilize Graph Neural Networks (GNNs) to model the underlying graph structure of traffic data~\cite{Yu2018ijcaip505,Wu2019ijcaip264,bai2020adaptive}. 
Additionally, Spatio-Temporal Neural Networks (STNNs) based on self-attention mechanisms have emerged learning spatial and temporal relationships in a data-driven manner without prior knowledge~\cite{Zheng_Fan_Wang_Qi_2020,park2020st,Jiang_Han_Zhao_Wang_2023}.\par

\begin{figure*}
    \centering
        \includegraphics[width=1\linewidth]{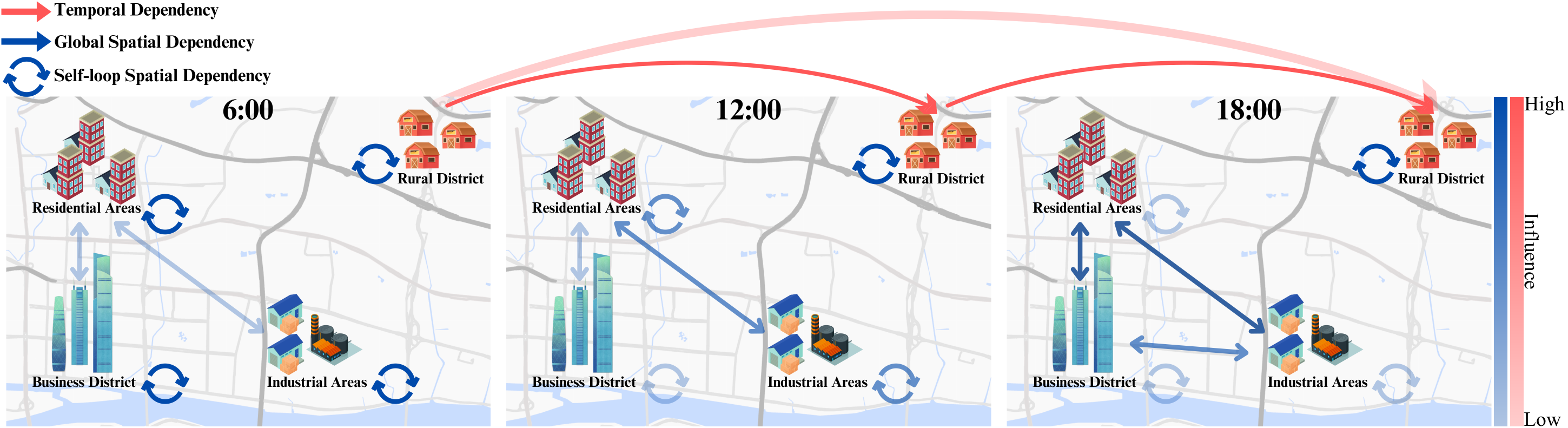}
    \caption{An example of a traffic flow system at three different times.}
    \label{fig:1}
\end{figure*}

As typical time series data, traffic data exhibit significant temporal dependency, as well as unique time-varying spatial relationships, which we refer to as ST dynamics.
%
As shown in Figure \ref{fig:1}, temporal dependency is a prominent characteristic in the traffic flow, with closer time points exhibiting stronger dependencies.
On the other hand, spatial dependencies are continuously changing over time.  
%
Therefore, modeling both temporal dependencies and ST dynamics is crucial for accurate traffic forecasting. Previous DL-based approaches tend to model temporal and spatial information separately in an attempt to capture temporal and spatial dependencies~\cite{Wu2019ijcaip264}. %
%
%
%
Moreover, in real-life scenarios, different functional areas at different times tend to have varying proportions of self-loop and adjacency relationship in graph structure, i.e., varying local and global information demands. Previous DL methods often overlook the varying needs of nodes for global and local information in different ST scenarios, failing to capture the underlying dynamic ST characteristics of the traffic time-series data effectively.
%
%

To overcome these limitations, we introduce a Dyn-ST embedding to simulate time-varying spatial relationship in real-life traffic data and  propose a Dynamic Spatio-Temporal Graph Transformer Network (DST-GTN) to dig out complex ST dynamics, based on the idea that ST dynamics are essentially superimposed on the temporal dimension.
%
%
The main contributions of this work can be summarized as follows:

\begin{itemize}
    \item We introduce a novel in-depth feature representation named Dynamic ST (Dyn-ST) features to simulate hidden ST dynamics in traffic time-series data and encapsulating time-varying spatial relationships.
    \item We propose a new traffic forecasting model named Dynamic Spatio-Temporal Graph Transformer Network (DST-GTN) to perform traffic forecasting accurately. Based on the idea that ST dynamics are superimposed on temporal dependencies, the DST-GTN can fully capture the complex ST dynamics in traffic time-series data. As the result, our method have a better performance.
    \item Extensive results show that our DST-GTN method achieves state-of-the-art (SOTA) performance and demonstrates competitive computational efficiency and robustness.
\end{itemize}


%

The remainder of this study is organized as follows: Section \ref{sec:related} presents related work on traffic forecasting. Fundamental concepts are investigated in Section \ref{Sec:Preliminary}. Section \ref{Sec:methods} introduces the details of proposed DST-GTN method. Section \ref{Sec:experiments} conducts experiments across diverse real-world datasets. Section \ref{Sec:Conclud} concludes the paper.

\section{Related Work}
\label{sec:related}
In this section, we summarize literature review related to traffic forecasting and introduce classic works of GNNs and Transformers and their applications in traffic forecasting.



\subsection{Traffic Forecasting}
%
Traditional work on traffic forecasting mostly falls into two categories: knowledge-driven and data-driven. 
Knowledge-driven approaches typically utilize queuing theory to simulate user behavior in traffic, but often fail to account for the complex nature of real-world traffic flow~\cite{cascetta2013transportation}. 
Data-driven methods are usually based on statistical methods. For instance, ARIMA~\cite{Williams2003JournalofTransportationEnginering} and VAR~\cite{lu2016integrating} are used to forecast the traffic condition of each traffic sensor separately.
However, these methods cannot handle the high non-linearity of each time series effectively. 
Furthermore, machine learning (ML) methods are proposed, such as support vector machine (SVM)~\cite{DruckerHarrisSupport_Vector_Regression_machines} and k-nearest neighbor (KNN)~\cite{van2012short}, which are effective in relaxing the linear dependency assumption. 
Although these models mainly consider temporal dependencies and tend to overlook spatial relationships between traffic sensors. 
To model both temporal and spatial dependencies, deep learning (DL) methods, including Recurrent Neural Networks (RNNs) and Convolutional Neural Networks (CNNs), have been increasingly popular among researchers.
In particular, these DL methods can extract latent representations and exploit much more features of traffic data than traditional ML methods~\cite{bui2022spatial}.
%
%
Nonetheless, they are not optimal for traffic forecasting because road networks possess graph structure.
%
%
Therefore, recent works in this field of study feature graph-based DL methods that capture the graph structure in traffic time-series data, such as 
STGNNs~\cite{Wu2019ijcaip264} and STNNs~\cite{Zheng_Fan_Wang_Qi_2020}.
%
%
%
%

\subsection{Graph Neural Network}
Graph Neural Networks (GNNs) are useful tools for learning spatial dependencies.
Common GNNs can be divided into three categories: spectral-based GCNs, spatial-based GCNs and GATs~\cite{jin2023spatio}. 
Spectral-based GCNs, as exemplified in~\cite{Bruna2014Learning_Representations}, introduce filters from graph signal processing in the spectral domain to define the graph convolution process. 
Among the spectral-based GCNs, ChebNet~\cite{Defferrard2016CNN_on_graphs_with_fast_localized_spectral_filtering} leverages a truncated expansion of Chebyshev polynomials up to the $k^{th}$ order to reduce the complexity of Laplacian computation. 
Furthermore, GCN~\cite{Thomas2017ICLR}, a first-order approximation of ChebNet, achieves outstanding performance in a variety of tasks. 
On the other hand, spatial-based GCNs, define the graph convolution process through information propagation. 
Diffusion graph convolution (DGC)~\cite{Atwood2016Diffusion_CNN} employs a transition probability from one node to its neighboring nodes to simulate graph convolution process.
GraphSAGE~\cite{hamilton2017inductive} uses sampling to obtain a fixed number of neighbors and models the graph convolutions as an information-passing process from one node to another directly connected node.
To dynamically learn the importance of neighbor nodes in learning spatial dependencies, GAT~\cite{veličković2018ICLR} integrates the attention mechanism into the node aggregation operation. 
Recently, GCNs has been adopted as a common component in STGNNs to model spatial relationships in ST data.
DCRNN~\cite{li2018diffusion} and STGCN~\cite{Yu2018ijcaip505} employed predefined graphs to capture spatial dependencies and adopt RNNs or CNNs to model temporal features.
%
%
Additionally, some studies, such as STSGCN~\cite{song2020spatial} and STFGNN~\cite{li2021spatial}, have attempted to simultaneously model temporal and spatial relationships by designing ST graphs.
%
\subsection{Transformer}
Transformer ~\cite{vaswani2017attention}, a self-attention based architecture, was initially applied to natural language processing (NLP) tasks and has achieved extraordinary performance across multiple NLP tasks~\cite{Devlin2019Human_Language_Technologies}.
In addition to its success in natural language processing, the Transformer architecture has demonstrated promising results in computer vision tasks, as evidenced by models like Vision Transformer~\cite{Bo2022CoRR} and Swin Transformer~\cite{liu2021swin}. 
%
%
%
%
%
In traffic forecasting, Transformer architecture based STNNs has also demonstrated outstanding competitive capabilities.
GMAN~\cite{Zheng_Fan_Wang_Qi_2020} employed an encoder-decoder architecture consisting of multiple ST attention blocks to model the impact of ST dynamics.
%
%
%
PDFormer~\cite{Jiang_Han_Zhao_Wang_2023} adopts multiple self-attention mechanisms with different design perspectives in its encoder to capture both temporal and spatial dependencies. 
Similarly, STAEformer~\cite{liu2023spatio} utilized the encoder of the Vanilla Transformer, applying it from both temporal and spatial perspectives.

\section{Preliminaries} 
\label{Sec:Preliminary}
In this section, we first introduce background knowledge of our work, including GCN and Transformer. Next, we define the traffic network, traffic time-series data and multi-step traffic forecasting problem addressed. 
\subsection{Preliminary of GCN}
Spectral-based GCNs~\cite{Bruna2014Learning_Representations} are a class of classical models of GCNs. Spectral-based GCNs, i.e. GCN~\cite{Thomas2017ICLR} offering a more efficient approach to graph convolution, formulate the propagation rule with
\begin{equation}
Z^{(l+1)} = \left(I+D^{-\frac{1}{2}}AD^{\frac{1}{2}}\right)Z^{(l)}\mathcal{\theta},
\end{equation}

where $I$ is the identity matrix and $A$ is the adjacency matrix, with $\mathcal{\theta}$ representing the learnable parameters. In a GCN, $D^{-\frac{1}{2}}AD^{\frac{1}{2}}$ can be considered a low-pass fitler to obtain global information of given graph, while $I$ serves as an all-pass filter to capture local information. AKGNN~\cite{Ju2022AAAI}
transforms the GCN into
\begin{equation}
Z^{(l+1)} = \left(\frac{2\lambda^{l}_{\text{max}}-2}{\lambda^{l}_{\text{max}}} I+\frac{2}{\lambda^{l}_{\text{max}}}D^{-\frac{1}{2}}AD^{\frac{1}{2}}\right)Z^{(l)}\mathcal{\theta},
\end{equation}
where $\lambda^{l}_{\text{max}}\in(1,+\infty)$ is a learnable scalar used to control the low-pass and all-pass filters. 
When $\lim \lambda^{l}_{\text{max}}\to 1$, the form is approximated by $Z^{(l+1)}=\left(2D^{-\frac{1}{2}}AD^{\frac{1}{2}}\right)Z^{(l)}\mathcal{\theta}$, which represents a full low-pass filter and only considers global information. When $\lim \lambda^{l}_{\text{max}}\to \infty$, the form is approximated by $Z^{(l+1)}=(2I)Z^{(l)}\mathcal{\theta}$, which represents a full all-pass filter and only considers local information. 
\subsection{Preliminary of Transformer}
The Transformer described in ~\cite{vaswani2017attention} is composed of multiple Transformer blocks, each containing a self-attention component and a position-wise feed-forward network (FFN) component. After each component, residual connections and layer normalization are typically applied. Assuming that the input to the $l^{\text{th}}$ Transformer block is
$Z^{(l)}\in \mathbb{R}^{T^{T}\times  D}$, the self-attention process on $Z^{(l)}$ can be computed by
\begin{equation}
    Q = Z^{(l)}W^{Q},\; K = Z^{(l)}W^{K},\; V = Z^{(l)}W^{V}
\end{equation}
\begin{equation}
    \text{Attention}(Q,K,V) = \text{softmax}\left(\frac{Q{K^T}}{\sqrt{d_{K}}}\right)V
\end{equation}
where $W^{Q},W^{K}\in \mathbb{R}^{T^{D}\times  d_{K}},W^{V}\in \mathbb{R}^{T^{D}\times  d_{V}}$ are learnable parameters used to map $Z^{(l)}$ to queries, keys and values, respectively. 
%
\subsection{Problem Definition}

\begin{definition} [Traffic Network] A traffic network can be represented as either a directed or undirected graph $G = (V, E, A)$, where $V$ represents a set of traffic sensor nodes with a total of $N$ nodes ($|V| = N$), $E$ is a set of edges between these nodes, and $A \in \mathbb{R}^{N \times N}$ is an adjacency matrix indicating the reachability between sensor nodes. 
\\
\end{definition}
\begin{definition} [Traffic Time-Series Data] We denote $X_{t}\in \mathbb{R}^{N\times C}$ as traffic time-series data collected from $N$ traffic sensor nodes at time $t$, where traffic time-series data can represent either traffic flow or traffic speed. In our study, $C$ equals 1.\\
\end{definition}
\begin{definition} [Multi-step Traffic Forecasting] Given a historical traffic data sequence $\mathcal{X}=(X_{t-T},X_{t-T+1},\cdots,X_{t})\in \mathbb{R}^{T\times N\times C}$, our aim is to predict the future traffic data sequence $[X_{t+1},X_{t+2},\cdots,X_{t+T}]\in \mathbb{R}^{T\times N\times C}$ based on the observed historical values, i.e. multi-step traffic forecasting. We formulate the problem as finding a mapping function $f$ to forecast the next $T$ steps traffic time-series data based on the previous $T$ steps traffic time-series data, that is
\begin{equation}
[X_{t+1},X_{t+2},\cdots,X_{t+T}] = f(X_{t-T+1},X_{t-T}\cdots,X_{t}).
\end{equation}
\end{definition}

\section{Methods}
\label{Sec:methods}
In this section, we introduce our Dynamic Spatio-Temporal Graph Transformer Network (DST-GTN) model. 
We describe its architecture and the details of DST-GTN.



\subsection{Architecture of DST-GTN}
\begin{figure*}
    \centering
            \includegraphics[width=1\linewidth]{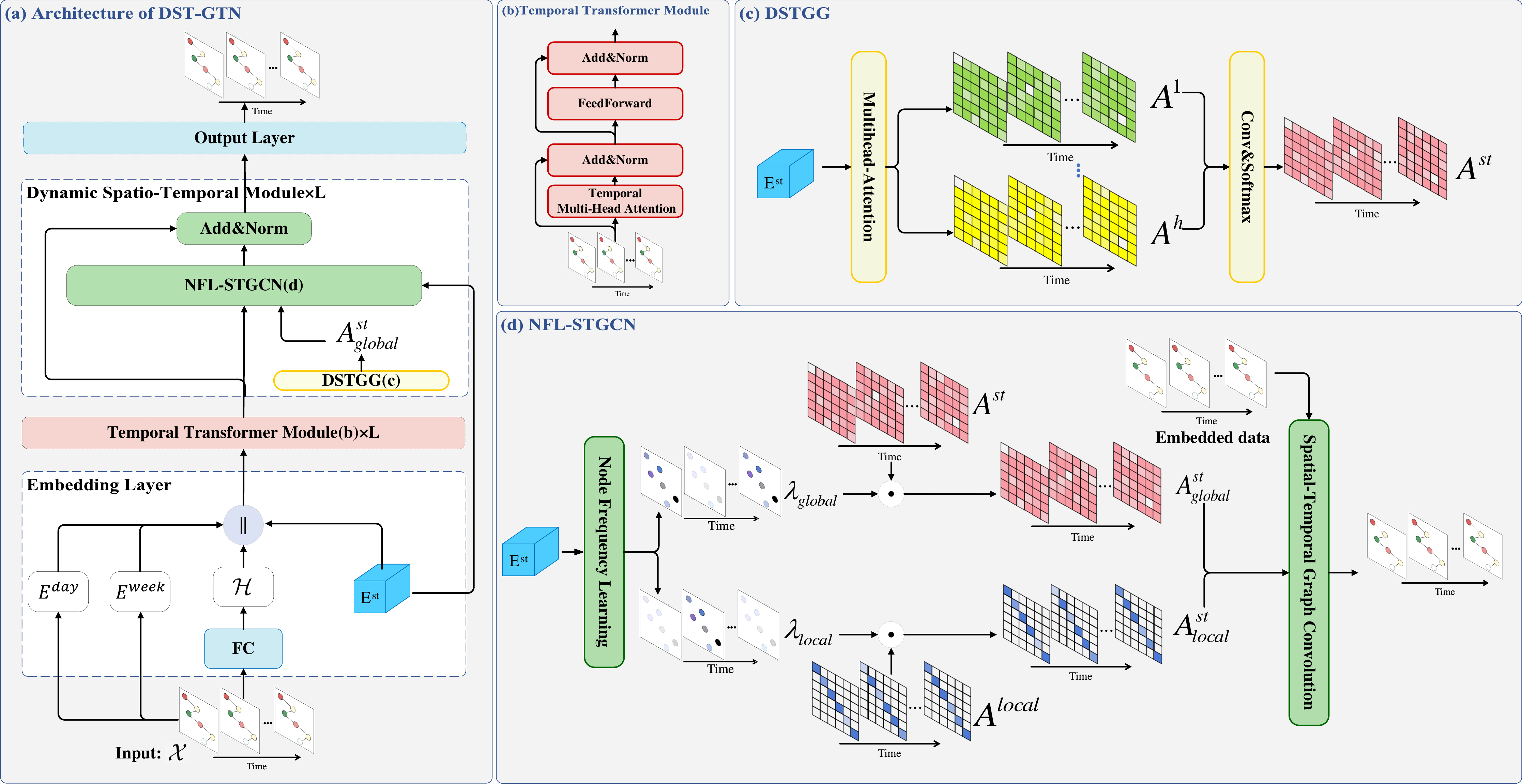}
    \caption{The detailed architecture of DST-GTN. (a) represents the overall architecture of DST-GTN. (b) describes the Temporal Transformer Module, which captures temporal dependencies in the embedded data. (c) illustrates the Dynamic Spatio-Temporal Graph Generator (DSTGG) component of the Dynamic Spatio-Temporal Module (DSTM), which utilize Dyn-ST embedding to generate a global ST graph. (d) is the NFL-STGCN component of the DSTM, which adaptively learns the local and global information demand for each ST node to adjust the ST graph. The optimized two ST graph is finally employed in the ST graph convolution process.}
    \label{fig:2}
\end{figure*}

We introduce the architecture of proposed Dynamic Spatio-Temporal Graph Transformer Network (DST-GTN) in Figure \ref{fig:2}. 
Specifically, The four main modules of the frame are summarized as follows:

\begin{itemize}

\item \textbf{Embedding Layer}: it generates traffic data embedding, temporal identity embedding and Dyn-ST embedding.
\item \textbf{Temporal Transformer Module}: it captures different hidden temporal dependencies along the time dimension. 
\item \textbf{Dynamic Spatio-Temporal Module}: it comprises two components: the Dynamic Spatio-Temporal Graph Generator (DSTGG) and the Node Frequency Learning Spatio-temporal Graph Convolution Network (NFL-STGCN). The DSTGG captures time-varying spatial relationships represented in Dyn-ST embedding. The NFL-STGCN component learns the global and local information demands of nodes in different ST scenarios. These two components jointly address the limitation mentioned in Introduction.
\item \textbf{Output Layer}: it forecasts all time slices for all nodes simultaneously utilizing the ST information generated by the past three parts.
\end{itemize}
\par

%

\subsection{Modules of DST-GTN}
 We detail about each module of DST-GTN.
\subsubsection{Embedding Layer}
We introduce two types of embedding in Embedding Layer: Temporal Identity Embedding and Dynamic Spatio-Temporal (Dyn-ST) Embedding, to jointly represent complex relationship of traffic data.\par
\begin{itemize}
    \item \textbf{Temporal Identity Embedding.} In order to enhance temporal distinguishability, we design two additional temporal identity embeddings $E^{week}\in \mathbb{R}^{T\times  d_{1}}$ and $E^{day}\in \mathbb{R}^{T\times  d_{1}}$ to jointly identify the traffic characteristics at a certain moment. Specifically, $E^{week}$ captures the temporal characteristics related to weekly periodicity, while $E^{day}$ focuses on daily periodicity. The detailed generation process of them is described in Appendix A.1.

\item \textbf{Dynamic Spatio-Temporal Embedding.} We introduce a dynamic spatio-temporal (Dyn-ST) embedding, denoted as $E^{st}\in \mathbb{R}^{N\times T\times  d_{2}}$, to simulate the complex ST dynamics in traffic data. Here $d_{2}$ is the embedding dimension of $E^{st}$. Note that $E^{st}$ is a randomly initialized trainable tensor, which is progressively updated during the training process to accurately represent the ST dynamics.\par
\end{itemize}
Raw traffic time-series data $\mathcal{X}$ are transformed into a high-dimensional representation $\mathcal{H}\in \mathbb{R}^{T\times N\times d}$ through a fully connected layer, where $d$ is the embedding dimension. The output of embedding layer is generated by concatenating the traffic data embedding, the temporal identity embedding, and the Dyn-ST embedding, as shown in the equation:

\begin{equation}
    \mathcal{Z} = \mathcal{H}\|E^{day}\|E^{week}\|E^{st},
\end{equation}
where $\mathcal{Z}\in \mathbb{R}^{N\times T\times  D}$ and $D=d+2d_{1}+d_{2}$. This output is then used as the input for the following Temporal Transformer module.

\subsubsection{Temporal Transformer Module}
To capture global temporal dependencies of embedded traffic data, we adopt Transformer blocks in Vanilla Transformer~\cite{vaswani2017attention} to form Temporal Transformer module. Each Temporal Transformer module contains two components: multi-head temporal self-attention (MTSA) and a position-wise feed-forward network (FFN).  Define the representation of the embedding data $\mathcal{Z}$ at the $i^{th}$ traffic sensor as $Z^{i}\in \mathbb{R}^{T\times D},i\in\{1,\cdots,N\}$. In MTSA component, $Z^{i}$ is firstly converted to query matrices $Q^{i}\in \mathbb{R}^{T\times d_{K}}$, key matrices $K^{i}\in \mathbb{R}^{T\times d_{K}}$, and value matrices $V^{i}\in \mathbb{R}^{T\times d_{V}}$ by matrix multiplication. The temporal self-attention (TSA) process on node $i$ can be calculated as:
\begin{equation}
    Q^{i} = Z^{i}W^{Q^{i}}, K^{i} = Z^{i}W^{K^{i}}, V^{i} = Z^{i}W^{V^{i}}
\end{equation}
\begin{equation}
    TSA_{i}(Q^{i},K^{i},V^{i}) = softmax(\frac{Q^{i}{K^{i}}^\top}{\sqrt{d_{K}}})V^{i}
\end{equation}

where $W^{Q}\in \mathbb{R}^{D\times  d_{K}}$, $W^{K}\in \mathbb{R}^{D\times  d_{K}}$, $W^{V}\in \mathbb{R}^{D\times  d_{V}}$ are learnable parameter. For simplicity, we set $d_{K} = d_{V} = D$. Due to the different temporal dependencies of traffic data sequences, we expand TSA to multi-head version with $h$ head, i.e., MTSA. Query, key, and value  are split into $h$ heads in the embedding dimension, which can be denoted as $Q^{i}_{m},K^{i}_{m},K^{i}_{m}\in \mathbb{R}^{T\times \frac{D}{h}}, m\in\{1,\cdots,h\}$. The calculation process of MTSA on node $i$ is denoted as:
\begin{gather}
    MTSA_{i}(Q^{i},K^{i},V^{i}) = W^{O}*Concat(head_{1},\cdots,head_{h})\\
    where\ head_{m} = TSA(Q^{i}_{m},K^{i}_{m},V^{i}_{m}) \nonumber
\end{gather}
We employ MTSA in parallel for all nodes on $\mathcal{Z}$ and concatenate all outputs in the node dimension. Therefore, the MTSA process on $\mathcal{Z}$ can be represented as:
\begin{gather}
    MTSA(\mathcal{Z}) = Concat(MTSA_{1}(Q^{1},K^{1},V^{1})),\cdots,\\
    \qquad \qquad  \qquad \ \ \ \ \ \  MTSA_{N}(Q^{N},K^{N},V^{N}))\nonumber
\end{gather}\par 
The FFN is implemented by multiple fully connected layers and closely follows MTSA's usage. Following vanilla Transformer, we adopt layer normalization (LayerNorm) after each MTSA component and FFN component. The overall process of $l^{th}$ Temporal Transformer module can be calculated as:
\begin{equation}
    \hat{\mathcal{Z}}^{(l)} = LayerNorm(MTSA(\mathcal{Z}^{(l-1)}) + \mathcal{Z}^{(l-1)})
\end{equation}
\begin{equation}
    \mathcal{Z}^{(l)} = LayerNorm(FFN(\hat{\mathcal{Z}}^{(l)}) + \hat{\mathcal{Z}}^{(l)})
\end{equation}
The output of the last Temporal Transformer block is denoted as $\mathcal{Z}$ as before for simplicity, which will be fed into DSTM.
\subsubsection{Dynamic Spatio-Temporal Module}
In the Dynamic Spatio-Temporal Module (DSTM), a DSTGG component utilize Dyn-ST embedding to construct a global ST graph. 
Following this, a NFL-STGCN component learn the demand of each ST nodes for local and global information. 
This learned information is used to refine the global and local ST graphs accordingly.
The optimized two ST graph is finally employed in the ST graph convolution process.
\begin{itemize}
    \item \textbf{DSTGG.} Considering Dyn-ST embedding $E^{st}\in \mathbb{R}^{T\times N\times d_{2}}$ can simulate the ST dynamics of real-world traffic time-series data, we can employ $E^{st}$ to capture the time-varying spatial relationships of traffic data, which can be represented by a ST graph. First, denoting the representation of $E^{st}$ at time $t$ is $E^{st}_{t}\in \mathbb{R}^{N\times d_{2}}$, the global spatial relationship of sensor nodes at time $t$ can be calculated using spatial self-attention (SSA) on $E^{st}_{t}$
\begin{equation}
    Q_{t} = E^{st}_{t}W^{Q_{t}}, K_{t} = E^{st}_{t}W^{K_{t}}
\end{equation}
\begin{equation}
    SSA_{t}(Q_{t},K_{t}) = \frac{Q_{t}{K_{t}}^\top}{\sqrt{d_{2}}}
\end{equation}
where $Q_{t},K_{t} \in\mathbb{R}^{N\times d_{k}}$ are query and key matrix of at time $t$ and $W^{Q_{t}},W^{K_{t}}\in\mathbb{R}^{d_{2}\times d_{k}}, d_{k}=d_{2}$ are learnable parameters. Note that different with the TSA component in Temporal Transformer module, here we only calculate the unnormalized attention matrix to express the spatial relationship. However, one spatial relationship is not enough. Spatial relationships in traffic data often exhibit multiple patterns, such as distance relationships and functional area relationships. To explicitly model different spatial patterns, we expand the aforementioned SSA into $h$ heads, with each head representing a possible spatial relationship:
\begin{equation}
    A^{1}_{t},\cdots,A^{h}_{t} = SSA_{1}(Q_{t}^{1},K_{t}^{1}),\cdots,SSA_{h}(Q_{t}^{1},K_{t}^{1})
\end{equation}
where $Q_{t}^{m},K_{t}^{m}\in \mathbb{R}^{N\times \frac{d_{2}}{h}}, m\in\{1,\cdots,h\}$ is the query, key of $m^{th}$ head, and $A_{t}^{m}\in\mathbb{R}^{N\times N}$ represent attention matrix of $m^{th}$ spatial pattern at time $t$. In this way, we can compute $h$ different attention matrix $\{A^{1}_{t},\cdots,A^{h}_{t}\}$ to represent $h$ different spatial pattern at time $t$. Then, we integrate the information of $h$ spatial attention matrix through convolution operation and normalize it to generate a  normalized adjacency matrix at time $t$:
\begin{equation}
    A_{t} = softmax(Conv(A^{1}_{t},\cdots,A^{h}_{t}))
\end{equation}
where the convolution operation uses $h*1$ convolution kernels to simultaneously mix the information of $h$ attention matrices.\par

We perform the aforementioned calculations in parallel at all time points to generate $T$ global spatial graph, and finally concatenate them along the time dimension to form a global ST graph $A^{st}\in \mathbb{R}^{T\times N\times N}$ that represents global ST relationships. 
\begin{equation}
     A^{st} = (A_{1}\|,\cdots,\|A_{T})
\end{equation}
The entire process of generating a global ST graph $A^{st}$ is defined as Dynamic Spatio-Temporal Graph Generator (DSTGG). 
\item \textbf{NFL-STGCN.}  In traffic data, different traffic sensor nodes often have different demands for global and local information. For instance, suburban nodes, which interact less with the outside, are more focused on local information, while those in urban centers, having frequent interactions with neighboring nodes, require a greater emphasis on global information. In addition, this characteristic tends to change over time.\par 
%
%
To model these characteristics, inspired by AKGNN~\cite{Ju2022AAAI}, we can assign different all-pass filter weights and low-pass filter weights for each node during graph convolution process, tailored to their specific ST contexts.\par

Specifically, we first utilize Dyn-ST embedding $E^{st}$ to 
learn expected all-pass filter weights and low-pass filter weights for each node, namely Node Frequency Learning (NFL):
\begin{equation}
    \lambda = 1 + ReLU(MLP(E^{st}))
\end{equation}
\begin{equation}
    \lambda_{local} = \frac{2\lambda-2}{\lambda},\lambda_{global} = \frac{2}{\lambda}
\end{equation}
where $\lambda,\lambda_{local},\lambda_{global}\in \mathbb{R}^{T\times N}$, and MLP is 2-layer fully connected layer with $ReLU$ activation. $\lambda_{local},\lambda_{global}$ indicate the expected all-pass and low-pass filter weights of each sensor nodes, respectively, which is uncovered in a data driving way.  Note that the sum of $\lambda_{local}$ and $\lambda_{global}$ is 2. When $\lambda$ is greater than 1, the model expect all-pass filter can dominate. When $\lambda$ is less than 1, it expect low-pass filter takes precedence.\par
After learn expected all-pass filter weights and low-pass filter weights for each node, we denote a local ST graph $A^{local}\in \mathbb{R}^{T\times N\times N}$ as a all-pass filter, which is generated by concatenating $T$ identity matrix with shape $N\times N$. The global ST graph $A^{st}$ generated by previous DSTGG can be considered as a low-pass filter.
We can use expected all-pass filter weights and expected low-pass filter weights to adjust these two filter:
\begin{equation}
    A^{st}_{local} = \lambda_{local}\odot A^{local},A^{st}_{global} = \lambda_{global}\odot A^{st}
\end{equation}
where $\odot$ represent element-wise multiplication and the missing dimensions of $\lambda_{local},\lambda_{global}$ will be broadcasted when element-wise multiplication\par 
Then, defining the representations of $A^{st}_{local},A^{st}_{global}$ generated at the $l^{th}$ layer as $A^{st\ (l)}_{local},A^{st\ (l)}_{global}$, we can expand spectral graph convolution to ST graph convolution to aggregate ST information of traffic data:
\begin{equation}
    \hat{\mathcal{Z}}^{(l)} =  (A^{st\ (l)}_{local}+A^{st\ (l)}_{global})\times\mathcal{Z}^{(l-1)}W
\end{equation}
where $\times$ represents tensor multiplication and $W\in\mathbb{R}^{D\times D}$ is a learnable parameter. $\mathcal{Z}^{(l-1)}\in\mathbb{R}^{T\times N\times D}$ is the output of ${l-1}^{th}$ layer and  $\hat{\mathcal{Z}}^{(l)}\in\mathbb{R}^{T\times N\times D}$ is the ST convolution result in $l^{th}$ layer. The tensor multiplication can be viewed as performing graph convolution in parallel on different time slices. \par 
\end{itemize}
We stack $L$ DSTM.  In $l^{th}$ DSTM, assumping the output of NFL-STGCN is $\hat{\mathcal{Z}}^{(l)}$, we employ residual connection and layer normalization to generate the final output $\mathcal{Z}^{(l)}$ of $l^{th}$ DSTM:
\begin{equation}
    \mathcal{Z}^{(l)} = LayerNorm(\hat{\mathcal{Z}}^{(l)} + \mathcal{Z}^{(l-1)})
\end{equation}

\subsubsection{Output Layer}
Finally, we directly employ a MLP to transform the output of last DSTM to the expected forecasting sequence $\mathcal{Y}\in \mathbb{R}^{T\times N\times C}$, which can be represented as:
\begin{equation}
    \mathcal{Y} = MLP(\mathcal{Z}^{(L)})
\end{equation}
where $\mathcal{Z}^{(L)}\in \mathbb{R}^{T\times N\times D}$ is the output of the last DSTM.

\section{Experiments}
\label{Sec:experiments}


In this section, we compared the proposed DST-GTN with state-of-the-art models on five real-world traffic datasets. 
We design comprehensive ablation studies to evaluate the impact of each module.
Finally, we analyze the computational efficiency and robustness of DST-GTN.


\subsection{Datasets}

\label{sec:dataset}

We assessed the performance of DST-GTN using five publicly available real-word public traffic datasets: PEMS04, PEMS07, PEMS08, PEMS07(M) and PEMS07(L), which were collected from the Caltrans Performance Measurement System (PeMS)~\cite{Chen2001PeMS} . Among them, PEMS04, PEMS07, PEMS08 are traffic flow datasets~\cite{song2020spatial}, while PEMS07(M), PEMS07(L) are traffic speed datasets. The statistical details of these five datasets are summarized in Table \ref{tab:booktabs}.
\begin{table}[H]
\begin{center}
\resizebox{240pt}{!}{%
\begin{tabular}{*{5}{lllrr}}
    \toprule[0.4mm]
 Datasets  &Type &Time Interval & Node & Time Steps  \\	
 \midrule[0.5mm]
PEMS04    & Flow &5 min   & 307  & 16992   \\
PEMS07    & Flow &5 min  & 883  & 28224     \\
PEMS08    & Flow  &5 min & 170  & 17856     \\
PEMS07(M) & Speed &5 min & 228  & 12672     \\
PEMS07(L) & Speed &5 min & 1026 & 12672  \\
  \bottomrule[0.5mm]
  \end{tabular}}
  \end{center}
  \caption{Summary statistics of five traffic datasets.}
  \label{tab:booktabs}
\end{table}
\subsection{Baselines}
\label{sec:baseline}
We compared our model with the following 12 baselines: VAR~\cite{lu2016integrating}, STGCN~\cite{Yu2018ijcaip505}, DCRNN~\cite{li2018diffusion}, ASTGCN(r)~\cite{Guo_Lin_Feng_Song_Wan_2019}, GWNET~\cite{Wu2019ijcaip264}, STSGCN~\cite{song2020spatial}, AGCRN~\cite{bai2020adaptive}, STFGNN~\cite{li2021spatial}, STGNCDE~\cite{choi2022graph}, STID~\cite{shao2022spatial}, PDFormer~\cite{Jiang_Han_Zhao_Wang_2023}, and STAEformer~\cite{liu2023spatio}. 

\subsection{Performance Comparison}
  \begin{table*}
\begin{center}
	\resizebox{\textwidth}{!}{%
\begin{tabular}{c|c|cccccccccccc|c}
\toprule[0.4mm]
          & Models & VAR   & STGCN & DCRNN & ASTGCN(r) & GWNET & STSGCN & AGCRN & STFGNN & STGNCDE & STID        & PDFormer & STAEformer  & DST-GTN        \\
            \midrule[0.2mm]
         \multirow{3}{*}{PEMS04}  & RMSE   & 36.39 & 34.77 & 31.43 & 35.22     & 31.72         & 33.65  & 31.25 & 31.87  & 31.09   & \underline{29.93} & 30.05    & 30.20       & \textbf{29.91} \\
             & MAE    & 23.51 & 21.76 & 19.71 & 22.92     & 19.36         & 21.19  & 19.38 & 19.83  & 19.21   & 18.41       & 18.39    & \underline{18.24} & \textbf{18.12} \\

          & MAPE (\%)   & 17.85 & 13.87 & 13.54 & 16.56     & 13.30         & 13.90  & 13.40 & 13.02  & 12.77   & 12.59       & 12.15    & \underline{12.04} & \textbf{11.91} \\
           \midrule[0.2mm]
       \multirow{3}{*}{PEMS07}   & RMSE   & 55.73 & 35.44 & 34.43 & 37.87     & 34.12         & 39.03  & 34.40 & 35.81  & 34.04   & 32.74       & 32.87    & \underline{32.60} & \textbf{32.40} \\
  & MAE    & 37.06 & 22.90 & 21.20 & 24.01     & 21.22         & 24.26  & 20.57 & 22.07  & 20.62   & 19.63       & 19.83    & \underline{19.14} & \textbf{19.00} \\
          & MAPE (\%)   & 19.91 & 11.98 & 9.06  & 10.73     & 9.08          & 10.21  & 8.74  & 9.21   & 8.86    & 8.31        & 8.52     & \underline{8.01}  & \textbf{7.95}  \\
           \midrule[0.2mm]
  \multirow{3}{*}{PEMS08}        & RMSE   & 31.02 & 27.12 & 24.28 & 28.06     & 24.86         & 26.80  & 24.41 & 26.21  & 24.81   & 23.37       & 23.52    & \underline{23.37} & \textbf{22.91} \\
   & MAE    & 22.07 & 17.84 & 15.26 & 18.25     & 15.06         & 17.13  & 15.32 & 16.64  & 15.46   & 14.21       & 13.69    & \underline{13.54} & \textbf{13.40} \\
          & MAPE (\%)   & 14.04 & 11.21 & 9.96  & 11.64     & 9.51          & 10.96  & 10.03 & 10.55  & 9.92    & 9.29        & 9.07     & \underline{8.89}  & \textbf{8.87}  \\
           \midrule[0.2mm]
 \multirow{3}{*}{PEMS07(M)}         & RMSE   & 7.61  & 6.79  & 7.18  & 6.18      & 6.24          & 5.93   & 5.54  & 5.79   & 5.39    & \underline{5.36}  & 5.75        & 5.54        & \textbf{5.27}  \\
 & MAE    & 4.25  & 3.86  & 3.83  & 3.14      & 3.19          & 3.01   & 2.79  & 2.90   & 2.68    & \underline{2.61}  & 2.86        & 2.68        & \textbf{2.59}  \\
          & MAPE (\%)   & 10.28 & 10.06 & 9.81  & 8.12      & 8.02          & 7.55   & 7.02  & 7.23   & 6.76    & \underline{6.63}  & 7.18        & 6.70        & \textbf{6.59}  \\
         \midrule[0.2mm]
    \multirow{3}{*}{PEMS07(L)}   & RMSE   & 8.09  & 6.83  & 8.33  & 6.81      & 7.09          & 6.88   & 5.92  & 5.91   & \underline{5.76}    & 5.78  & 5.90        & 5.93        & \textbf{5.69}              \\
& MAE    & 4.45  & 3.89  & 4.33  & 3.51      & 3.75          & 3.61   & 2.99  & 2.99   & 2.87    & \underline{2.81}  & 2.93        & 2.84        & \textbf{2.80}              \\
          & MAPE (\%)   & 11.62 & 10.09 & 11.41 & 9.24      & 9.41          & 9.13   & 7.59  & 7.69   & 7.31    & \underline{7.15}  & 7.37        & 7.19        & \textbf{7.09}   \\
          \bottomrule[0.5mm]
\end{tabular}}
\caption{Performance comparison of DST-GTN and other baseline models. The best results are in bold and underline denotes the second-best results.}
\label{tab:performance}
\end{center}
\end{table*}
Table \ref{tab:performance} shows the comparison of different models for
the 1-hour ahead multi-step traffic forecasting tasks. The best results are highlighted in bold, while the second-best results are underlined. It is evident that our DST-GTN model outperforms other state-of-the-art models in all cases.
These experimental results show that DST-GTN greatly improves the accuracy of traffic forecasting. The detailed experimental settings and the analysis of the advantages of DST-GTN are shown in Appendix B.2 and B.3. \par 

\subsection{Ablation Study}

To assess the actual performance of each
component in DST-GTN, we conducted comparisons between DST-GTN with 5 variants: w/o TT, w/o TT-ST, DST-GTN-Reverse, w/ Static Graph, w/o NFL. The definition of these variants and the ablation study results are shown in in Appendix B.4.\par

The results lead us to the following conclusions: 
(1) The removal of Temporal Transformer modules and DSTM results in significant performance degradation. This highlights the importance of both temporal dependency and ST dynamics in traffic forecasting.
(2) DST-GTN-Reverse performs worse than DST-GTN, suggesting that extracting ST dynamics should be based on the extraction of temporal dependencies. This structure where temporal dependencies are followed by ST dynamics aligns with the intrinsic characteristics of traffic forecasting.
(3) Replacing global ST graphs with static graphs leads to a performance decrease, indicating that static graphs are inadequate for modeling the spatial relationships at different sequence positions.
(4) W/o NFL performs worse than DST-GTN. This validates the effectiveness of the node frequency learning component, which can accurately identify the demands of different ST nodes for global and local information, thereby aiding in mining ST dynamics.

\subsection{Model Efficiency and Robustness} 
\begin{table}[H]
\begin{center}
\resizebox{240pt}{!}{%
\begin{tabular}{l|ll|ll}
\toprule[0.4mm]
Datasets   & \multicolumn{2}{c|}{PEMS04}               & \multicolumn{2}{c}{PEMS08}                \\
\midrule[0.2mm]
Models      & \multicolumn{1}{c|}{Inference} & Training & \multicolumn{1}{c|}{Inference} & Training \\
\midrule[0.2mm]
STGNCDE    & \multicolumn{1}{c|}{15.55}     & \multicolumn{1}{c|}{150.44}   & \multicolumn{1}{c|}{11.06}    & \multicolumn{1}{c}{107.26}   \\

PDFormer   & \multicolumn{1}{c|}{11.07}     & \multicolumn{1}{c|}{116.90}   & \multicolumn{1}{c|}{5.02}      & \multicolumn{1}{c}{61.11}    \\

STAEformer & \multicolumn{1}{c|}{6.77}      & \multicolumn{1}{c|}{66.10}    & \multicolumn{1}{c|}{3.64}      & \multicolumn{1}{c}{37.62}    \\

\textbf{DST-GTN}    & \multicolumn{1}{c|}{\textbf{4.31}}      & \multicolumn{1}{c|}{\textbf{40.19}}    & \multicolumn{1}{c|}{\textbf{2.52}}      & \multicolumn{1}{c}{\textbf{27.45}}     \\
\bottomrule[0.5mm]

\end{tabular}}
\caption{Training time (s/epoch) and inference time (s) on PEMS04 and PEMS08.}
\label{tab:time}
\end{center}
\end{table}

We compared the computational cost of DST-GTN with the self-attention-based models PDFormer and STAEformer, and the GCN-based model STGNCDE, on the PEMS04 and PEMS08 datasets. Table \ref{tab:time} displays the average inference time and the average training time per epoch. In comparison with these three baseline models, DST-GTN achieves a reduction in both inference and training times by at least 36.34\% and 39.20\%, respectively, on the PEMS04 dataset.

\begin{figure}[H]
    \centering
        \includegraphics[width=1\linewidth]{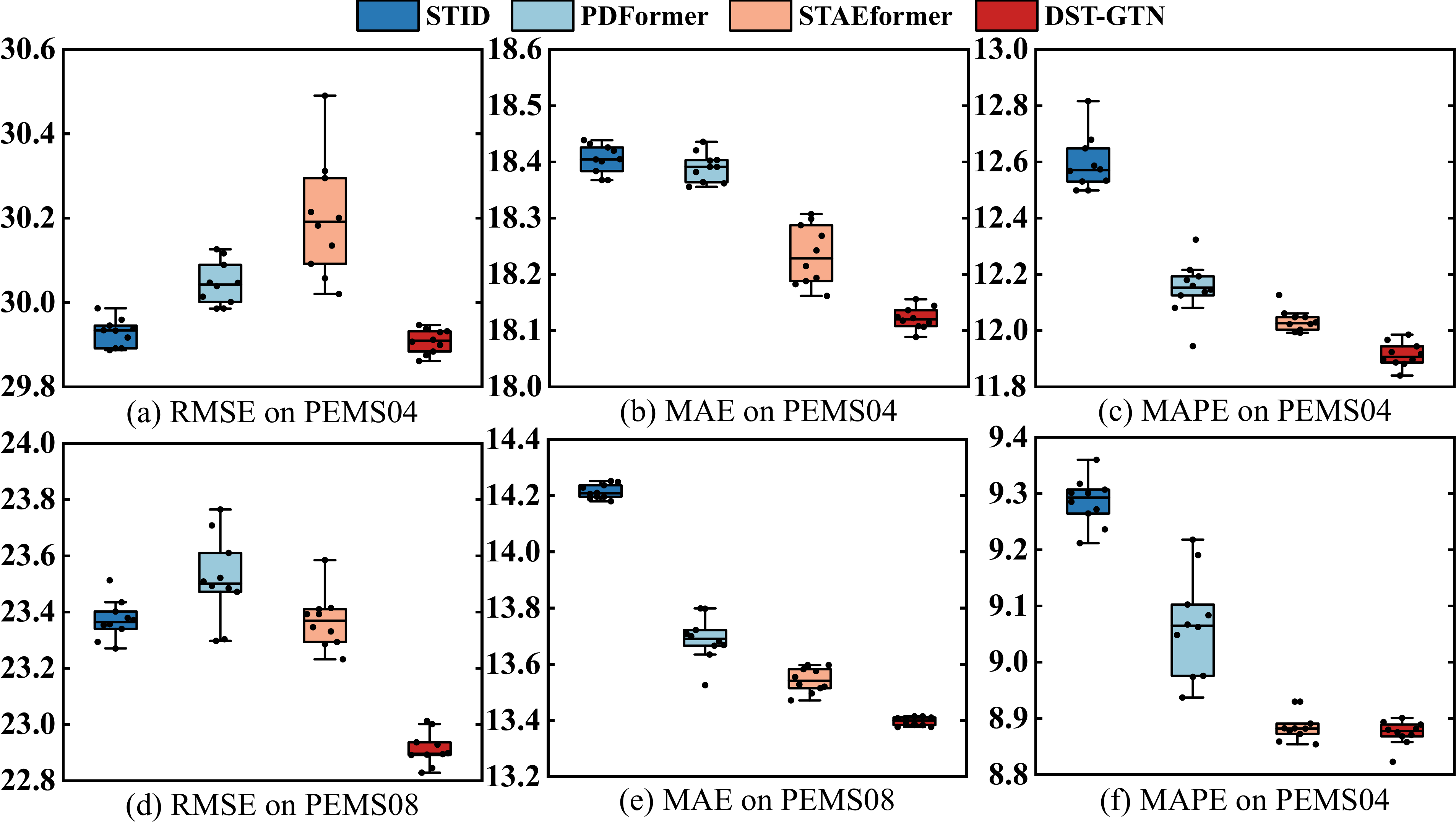}
    \caption{Robustness of different models on PEMS04 and PEMS08 datasets}
    \label{fig:3}
\end{figure}

Figure \ref{fig:3} illustrates the robustness between DST-GTN and the other three top-performing baselines on the PEMS04 and PEMS08 datasets, based on 10 experiments with different random seeds. 
As observed, DST-GTN typically has the shortest box in most cases, which means that DST-GTN has the best robustness.
\section{Conclusion}
\label{Sec:Conclud}
In this work, we proposed a Dynamic Spatio-Temporal Graph Transformer Network (DST-GTN) for traffic forecasting. 
%
%
Specifically, based on the idea that ST dynamics are superimposed on temporal dependencies, DST-GTN first employed a Temporal Transformer module to extract temporal dependencies. Then, DST-GTN used Dyn-ST embedding to simulate ST dynamics and employed DSTM to dig out the in-depth ST dependency in traffic data.
Extensive experiments on five real-world traffic datasets demonstrate that our model can outperform the state-of-the-art baselines.




\bibliographystyle{named}
\bibliography{ijcai23}
\clearpage
\appendix
\section{Details of DST-GTN}
\subsection{Temporal Identity Embedding.}
To obtain two temporal identities $E^{week}$ and $E^{day}$, we define a learnable parameter $I^{day}\in \mathbb{R}^{T^{d}\times  d_{1}}$ and another learnable parameter $I^{week}\in \mathbb{R}^{T^{w}\times  d_{1}}$ for identifying each sampling time within a day and each day within a week, respectively. Here, $T^{d}$ depends on the number of traffic data samples collected within a day, which typically is equal to 288, while $T^{w}$ is equal to 7. Subsequently, the original temporal information of each input traffic data will be mapping to corresponding daily periodicity representation among $I^{day}$ and weekly periodicity representation among $I^{week}$. The mapping result, $E^{week}$ and $E^{day}$, can be used to jointly identify temporal characteristics of each time slice of input traffic data.

\subsection{Algorithm of DST-GTN}
We summarize the overall computation process of DST-GTN as the following Algorithm 1. 
\begin{algorithm}[H]
	\renewcommand{\algorithmicrequire}{ \textbf{Input:}}     
\renewcommand{\algorithmicensure}{ \textbf{Output:}}    

	\caption{The overall algorithm of DST-GTN}
	\label{alg:1}
\begin{algorithmic}[1]
\REQUIRE The traffic data sequence over the past $T$ time steps $\mathcal{X}$ and their corresponding time stamp.
\ENSURE  The forecasting of future traffic data sequence $\mathcal{Y}$.
\STATE Initialize $E^{day}$ and $E^{week}$ based on time stamp.
\STATE Initialize $E^{st}$ randomly. 
\STATE $\mathcal{H}\leftarrow MLP(\mathcal{X})$

\STATE $\mathcal{Z}^{(0)}\leftarrow 
\mathcal{H}\|E^{day}\|E^{week}\|E^{st}$

\FOR{$l$ in range($L$)}
\STATE  $\hat{\mathcal{Z}}^{(l)} = LayerNorm(MTSA(\mathcal{Z}^{(l-1)}) + \mathcal{Z}^{(l-1)})$
\STATE  $\mathcal{Z}^{(l)} = LayerNorm(FFN(\hat{\mathcal{Z}}^{(l)}) + \hat{\mathcal{Z}}^{(l)})$
\ENDFOR

\STATE $\mathcal{Z}^{(0)}\leftarrow \mathcal{Z}^{(L)}$

\FOR{$l$ in range($L$)}
\STATE Calculate $A^{st(l)}_{local}$ and $A^{st(l)}_{global}$  according to Eq. 13 - Eq. 20.
\STATE $\hat{\mathcal{Z}}^{(l)} \leftarrow  (A^{st\ (l)}_{local}+A^{st\ (l)}_{global})\times\mathcal{Z}^{(l-1)}W$
\STATE    $\mathcal{Z}^{(l)} \leftarrow LayerNorm(\hat{\mathcal{Z}}^{(l)} + \mathcal{Z}^{(l-1)})$
\ENDFOR

\STATE $\mathcal{Y} \leftarrow MLP(\mathcal{Z}^{(L)})$
\end{algorithmic}  
\end{algorithm}

\section{Details of Experiment}

\subsection{Baseline}
\begin{itemize}
\item VAR is an classic time series method, which capture the pairwise correlations among time series.
\item STGCN utilizes temporal convolution network (TCN) and GCN to capture temporal dependencies and spatial dependencies in traffic data, respectively.
\item DCRNN captures the spatial dependency using bidirectional random walks on the graph, and the temporal dependency using the encoder-decoder architecture with scheduled sampling.
\item ASTGCN(r) combines an attention mechanism with graph convolution to model traffic data.
\item GWNET uses Gated TCN and GCN to model the temporal dependencies and spatial dependencies in traffic data, respectively, and introduces a self-adaptive adjacency matrix to learn spatial adjacency relationships.
\item STSGCN captures the complex localized spatial-temporal correlations through  spatial-temporal synchronous modeling mechanism.
\item AGCRN integrates GCN into the computational process of RNN to model time-varying spatial relationship.
\item STFGNN synchronously captures the ST correlation through the fusion of multiple spatial and temporal graphs.
\item STGNCDE extends the concept of neural controlled differential equations to temporal processing and spatial processing in traffic data. 
\item STID designs multiple embedding and use simple multi-layer perceptrons to forecast traffic data.
\item PDFormer utilizes multiple self-attention mechanism from spatial and temporal perspectives to achieve traffic forecasting.
\item STAEformer utilizes Vanilla Transformer from both temporal and spatial perspectives and introduces multiple embeddings.
\end{itemize}

\subsection{Settings}
\paragraph{Dataset Processing.} Consistent with previous research~\cite{song2020spatial,Weng2023Pattern_Recognition,Jiang_Han_Zhao_Wang_2023}, we divided all five datasets in chronological order, allocating 60\% for training, 20\% for validation, and 20\% for testing, following a 6:2:2 ratio. 
Previous 12 time steps traffic data were used to forecast the traffic data for the next 12 time steps, i.e., an equal-length multi-step forecasting. 
All datasets were normalized by Z-score normalization to standardize the inputs.\par
\paragraph{Hyper-parameters.} The embedding dimension of $E^{day}$,$E^{week}$, and $\mathcal{H}$ was set to 24. The embedding dimension of $E^{st}$ was set to 80. All number of multi-head attention heads were set to 4. The number of both Temporal Transformer module and DSTM were set to 3. 
We trained our model by using Adam optimzier and the training epoch was set to 200. The learning rate was set to 0.001 and the batch size was equal to 16. Additionally, early stopping strategy was used to avoid overfitting. We used the mean absolute error (MAE) as loss function to train our model.\par

\paragraph{Evaluation Metric.} We adopted three metrics to evaluate the performance of model: the first one is Mean Absolute Error (MAE), the second one is Root Mean Square Error (RMSE), and the third one is Mean Absolute Percentage Error (MAPE).
Missing values in the data were not be considered during the evaluation process.\par
\paragraph{Platform.} All experiments were conducted on a machine equipped with an NVIDIA GeForce 3090 GPU. DST-GTN was implemented using Pytorch 1.12.1 and Python 3.9.12. 
All experiments were repeated 10 times with different random seeds and the average results were reported.

\subsection{Analysis of Comparison Experiment}
VAR considers only the temporal dependency of traffic time-series data but ignores the crucial spatial information. 
Models like STGCN, ASTGCN(r), GWNET, STGNCDE address both temporal and spatial dependencies using
two separate modules, respectively. However, they overlook the dynamic ST interactions in traffic time-series data.\par 
To modeling ST dynamics, DCRNN and AGCRN integrate GNN with RNN to simulate time-varying spatial relationships. In contrast, STSGCN and STFGNN utilize pre-defined ST graphs to represent ST relationships over multiple steps.
Among these, DCRNN, STSGCN, and STFGNN employ pre-defined graph structures, potentially introducing incorrect prior knowledge into the models. AGCRN employs learnable node embedding to adaptively learn the graph structure and demonstrates better result. DST-GTN achieves better performance by learning ST dynamics through Dyn-ST embedding without relying on any prior knowledge.\par 
STID exhibits exceptional performance using only a MLP combined with temporal and spatial embedding. This proves the effectiveness of well-applied embedding. 
PDFormer and STAEforemr are entirely relied on self-attention and show promising result on traffic forecasting, which indicate the postive impact of self-attention mechanism.
However, both STID, PDFormer, and STAEformer overlook the precise modeling of ST dynamics. For instance, STAEformer applies learnable ST embedding to traffic data but does not fully exploit them.
In contrast,  DST-GTN employs the self-attention mechanism to capture the information from Dyn-ST embedding. The generated ST graph integrates both observed global and local information demands of each ST nodes to represent ST dynamics.\par 
Another key factor contributing to the superior performance of DST-GTN is its well-considered architectural design. 
The previously mentioned baseline models often concentrate on extracting spatial dependencies after temporal dependencies or on extracting both temporal and spatial dependencies. In contrast, DST-GTN goes a step further by extracting ST dynamics on top of temporal dependencies, which is intuitively more consistent with the characteristics of traffic time-series data.

\subsection{Details of Ablation Study}
\begin{table}[H]
\begin{center}
    
\resizebox{
240pt}{!}{%
\begin{tabular}{
l|ccc|ccc}

\toprule[0.4mm]
Datasets         & \multicolumn{3}{c|}{PEMS04}                      & \multicolumn{3}{c}{PEMS08}                                    \\
\midrule[0.2mm]
Models            & \multicolumn{1}{c}{RMSE}           & \multicolumn{1}{c}{MAE}            & \multicolumn{1}{c|}{MAPE(\%)}       & \multicolumn{1}{c}{RMSE}                         & \multicolumn{1}{c}{MAE}            & \multicolumn{1}{c}{MAPE(\%)}      \\
\midrule[0.2mm]
w/o TT           & 31.35          & 19.73          & 14.18          & 24.56                        & 15.43          & 10.86         \\
    
w/o TT-ST        & 39.89          & 25.17          & 18.00          & 31.49                        & 19.96          & 18.52         \\
DST-GTN Reverse  & 30.01          & 18.34          & 12.02          & 23.19                        & 13.74          & 9.09          \\
w/ Static Graph  & 30.05          & 18.34          & 12.10          & 23.06                        & 13.64          & 9.10          \\
w/o NFL          & 30.08          & 18.29          & 12.17          & 23.20 & 13.54          & 9.13          \\
\textbf{DST-GTN} & \textbf{29.91} & \textbf{18.12} & \textbf{11.91} & \textbf{22.91}               & \textbf{13.40} & \textbf{8.87}\\
\bottomrule[0.5mm]
\end{tabular}}
\caption{Ablation study on PEMS04 and PEMS08 datasets.}
\end{center}
 \label{table:A}
\end{table}

The definition of different variants in ablation study are described as follows: \par
(1)\textbf{w/o TT}: It removes Temporal Transformer module.\par
(2)\textbf{w/o TT-ST}: It removes Temporal Transformer module and  Dynamic Spatio-Temproal Module.\par
(3)\textbf{DST-GTN-Reverse}: It reverses the order of Temporal Transformer module and Dynamic Spatio-Temproal Module.\par
(4)\textbf{w/ Static Graph}: It replaces global  ST graph generate by DSTGG with static graph generated by distance between traffic nodes.\par
(5)\textbf{w/o NFL}: It removes the Node Frequency Learning component.\par
 Table A presents a comparison of these variants on the PEMS04 and PEMS08 datasets.

\section{Traffic Forecasting Visualization}
\begin{figure*}
    \centering
        \includegraphics[width=1\linewidth]{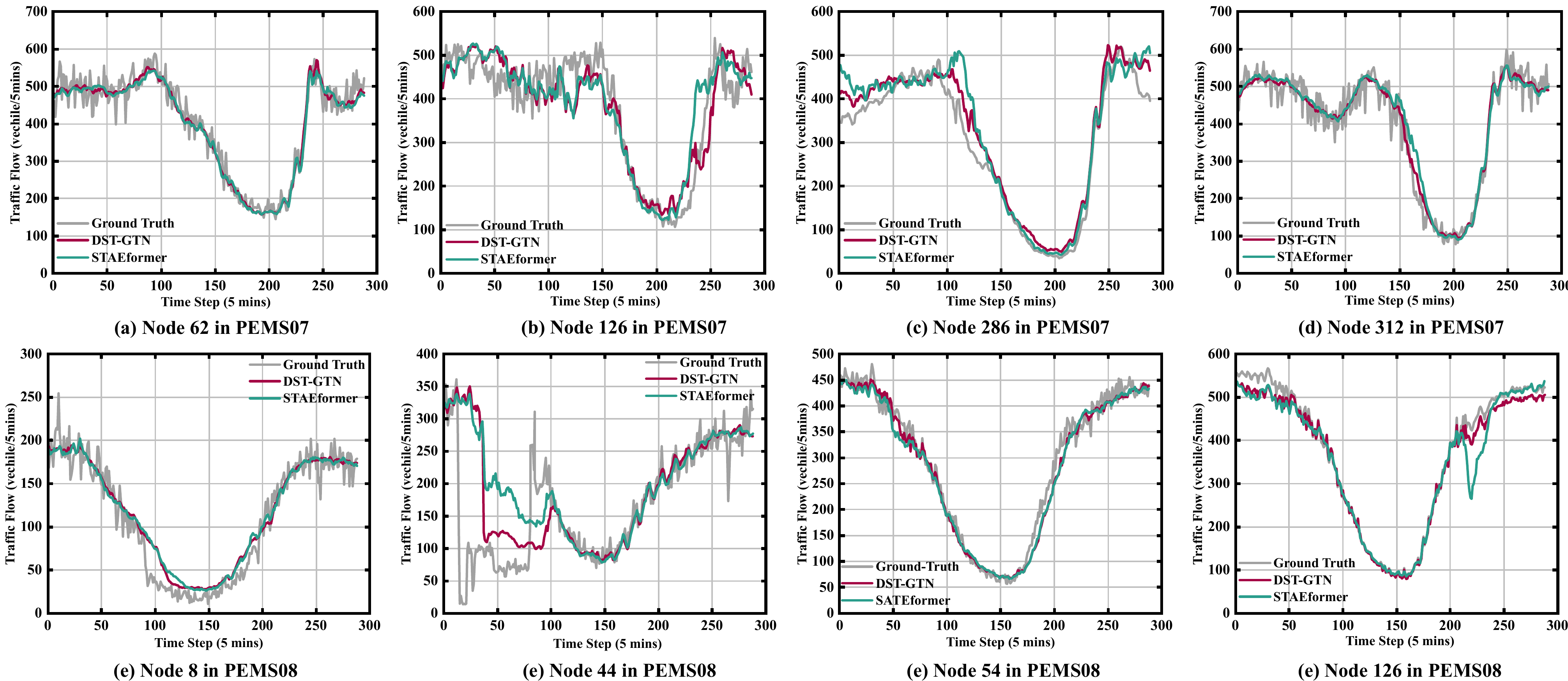}
    \caption{Traffic forecasting visualization.}
    \label{fig:A}
\end{figure*}

In this section, we visualize the ground truth and some forecasting outcomes by our method and STAEformer in Figure \ref{fig:A}.
\end{document}